\documentclass{article} 
\usepackage{iclr2021_conference,times}

\usepackage{amsmath,amsfonts,bm}

\def\eqref#1{equation~\ref{#1}}

\def\1{\bm{1}}

\DeclareMathAlphabet{\mathsfit}{\encodingdefault}{\sfdefault}{m}{sl}
\SetMathAlphabet{\mathsfit}{bold}{\encodingdefault}{\sfdefault}{bx}{n}

\usepackage{hyperref}
\usepackage{url}

\usepackage{graphicx}
\usepackage{amsmath}
\usepackage{amssymb}
\usepackage{listings}
\usepackage{todonotes}
\usepackage{tabularx} 
\usepackage{booktabs} 
\usepackage{siunitx}  

\title{Numerical Reasoning for Financial Reports}

\begin{document}


\maketitle

\begin{abstract}
Financial reports offer critical insights into a company's operations, yet their extensive length—typically spanning 30-40 pages—poses challenges for swift decision-making in dynamic markets. To address this, we leveraged fine-tuned Large Language Models (LLMs) to distill key indicators and operational metrics from these reports basis questions from the user. We devised a method \footnote{Code available at: \url{https://github.com/Abhi23run/CSE8803_DLT_Project/}}to locate critical data, and leverage the FinQA dataset to fine-tune both Llama-2-7B and T5 models for customized question answering. We achieved results comparable to baseline on the final numerical answer, a competitive accuracy in numerical reasoning and calculation.

\end{abstract}

\section{Introduction}
Analyzing financial reports serves as a powerful tool for various stakeholders to gain crucial insights into a company's performance and health \cite{gupta2021context}.
Investors rely on these reports to assess the company's profitability, growth potential, and risk levels, aiding their investment decisions. For management, these reports offer a window into operational efficiency, helping in strategic planning and identifying areas for improvement. Creditors and lenders use this data to gauge a company's ability to meet financial obligations and assess lending risks. Additionally, regulators and government entities rely on financial reports to ensure compliance with accounting standards and regulations, fostering transparency and accountability within the market. Ultimately, the meticulous scrutiny of financial reports equips stakeholders with vital information, enabling informed decision-making and fostering trust within the business ecosystem. As shown in \cite{ijfs10040104} and \cite{art23}, individual investors tend to favor companies that provide accessible and transparent disclosures when making investment decisions.

\begin{figure}[ht]
    \centering
    \includegraphics[width=0.77\linewidth]{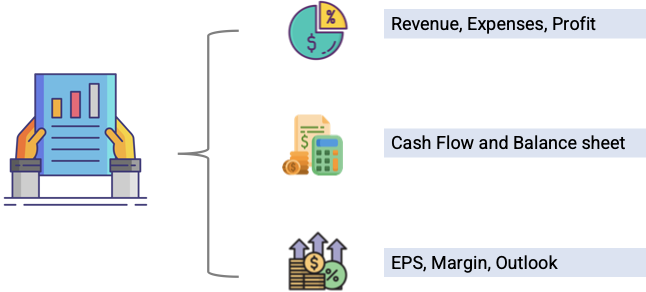}
    \caption{Annual reports help investors \& stakeholders to address the financial health of the firms}
    \label{fig:report_intoduction}
\end{figure}

Generally these reports offer information that falls into two broad categories: extractive data and numerical reasoning-based insights. Extractive data encompasses readily available information, such as revenue figures, expenses, and assets, which can be directly extracted from the reports. On the other hand, numerical reasoning involves analyzing trends, ratios, and financial indicators, requiring deeper numerical analysis to derive meaningful insights. Both types of information are crucial,\\ providing stakeholders with a comprehensive understanding of a company's financial performance and aiding decision-making processes.

\subsection{Related Works}
In the finance realm, there have been swift advancements in NLP \cite{gupta2023instruction,chung2022scaling}, especially in the finance domain. The extraction of information holds significant potential in finance, showcasing advantageous applications like sentiment analysis \cite{araci2019finbert}. In the past, research has ventured in three key directions. Initially, there were endeavors focused on mathematical calculations \cite{gupta2023john} within general domain question answering, as exemplified by datasets like DROP \cite{dua2019drop} and MaWPS\cite{koncel-kedziorski-etal-2016-mawps}. Secondly, there were studies emphasizing numerical reasoning concerning both table and text data, notable among them being HybridQA \cite{chen-etal-2020-hybridqa}. Lastly, there have been research efforts targeting NLP applications specifically in the financial domain, referenced as Financial NLP. However, an unaddressed gap existed—there was no prior work or dataset dedicated to constructing Question Answering (QA) systems focused on numerical reasoning derived from table data within financial reports. 
\subsection{Our contribution}
Within our project, our primary focus revolves around advancing research in two areas:
\begin{itemize}
    \item Firstly, our emphasis lies in enhancing numerical reasoning specifically tailored for financial reports, with a particular focus on interpreting and deriving insights from tabulated data. This endeavor aims to fortify the understanding and analysis of intricate financial information embedded within tables, allowing for nuanced and accurate comprehension.
    \item Secondly, we are developing an end-to-end pipeline that seamlessly extracts and generates insights directly from financial report PDFs. This comprehensive pipeline is designed to facilitate real-time analysis of financial reports, enabling swift and informed decision-making in dynamic market environments. By automating the extraction and interpretation of key data points, this system empowers users to swiftly access crucial insights, streamlining the process of financial analysis and enhancing overall efficiency.
\end{itemize}
   

\section{Dataset}
To accomplish our objective of numerical reasoning over financial reports, we leveraged an open source dataset - FinQA (\cite{chen2022finqa}) which is a large-scale dataset containing 8,281 examples written by financial experts, with fully annotated numerical reasoning programs. This dataset has been developed by leveraging the publicly available earnings reports of the S\&P 500 companies.

\begin{figure}[ht]
    \centering
    \includegraphics[width=1.0\linewidth]{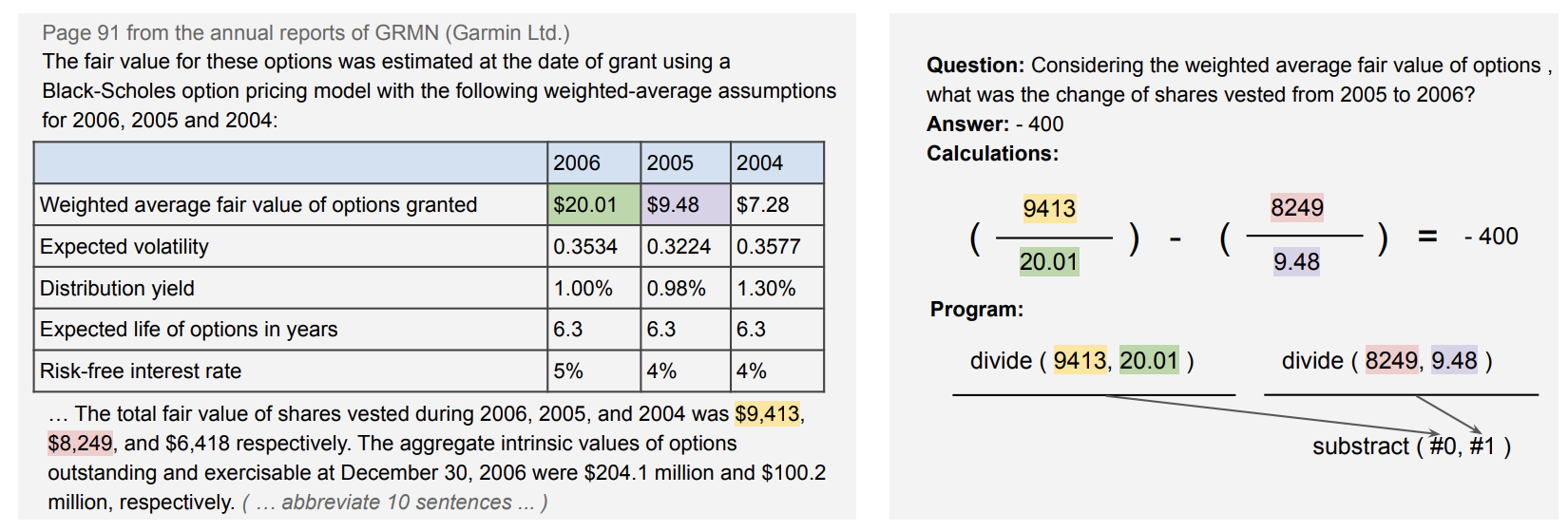}
    \caption{An example from FINQA: The model needs to learn how to calculate the number of shares, then select relevant numbers from both the table and the text to get the answer.}
    \label{fig:example_finqa_data}
\end{figure}

For the scope of our project, we have utilized the data with splits into training (6,053), validation (848) and test datasets (1,108), and contain mutually exclusive company reports. Additionally we have preprocessed the questions answers from FinQA into a step-wise structure as shown in \ref{fig:table_extraction_serialisation}.

\begin{figure}[ht]
    \centering
    \includegraphics[width=0.45\linewidth]{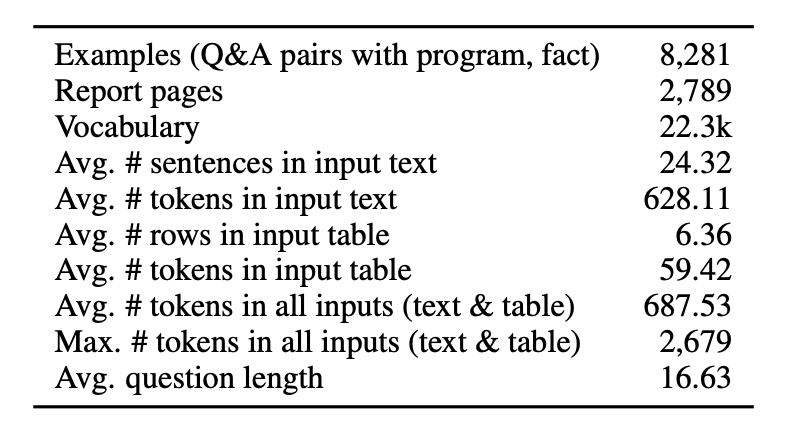}
    \includegraphics[width=0.45\linewidth]{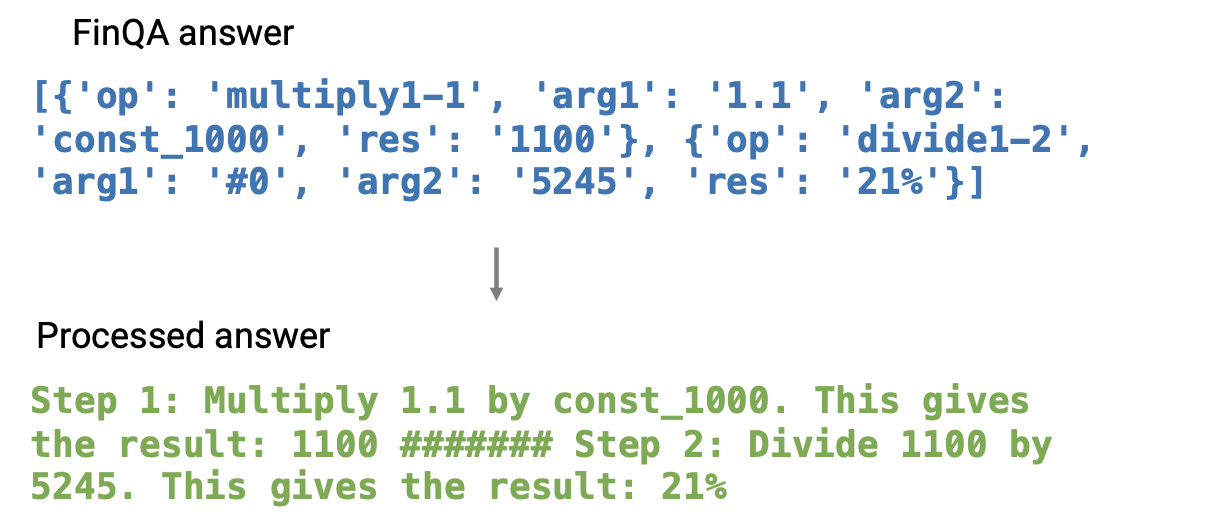}
    \caption{Statistics of FINQA (left) and Preprocessing for FinQA answers (right)}
    \label{fig:datatset_process}
\end{figure}

In the FINQA\footnote{FinQA Statistics taken from \cite{chen2022finqa}} dataset, answers to 23.42\% of the questions can be found solely within the textual information, while 62.43\% depend entirely on data from tables. A combination of both text and table data is necessary to answer 14.15\% of the questions. Regarding the facts presented, 46.30\% of the instances are based on a single sentence or table row; 42.63\% involve two pieces of information; and 11.07\% comprise more than two facts. For fine-tuning and training, the training dataset was utilized, and the validation dataset was employed for evaluation purposes.

\section{Methodology}

Our approach to addressing numerical reasoning questions rooted in financial reports involves a systematic sequence of steps. Initially, we employ a PDF parsing mechanism to extract tables embedded within the documents. These extracted tables are subsequently serialized into text, providing a structured and readable format for further processing. Following this, we generate vector embeddings of chunks from the serialized texts, enabling a numerical representation of the information contained within the tables. Leveraging these embeddings, we identify the most relevant context that aligns with the original question. Once the pertinent context is determined, we feed both the original question and the relevant context into a fine-tuned Language Model (LLM). 
This LLM, finetuned for financial report analysis, then produces accurate and context-aware answers to the numerical reasoning questions posed. 
This multi-step approach enhances the model's ability to comprehend and respond to complex queries based on financial data with precision. These different steps are discussed in subsequent sections below.

\begin{figure}[ht]
    \centering
    \includegraphics[width=1\linewidth]{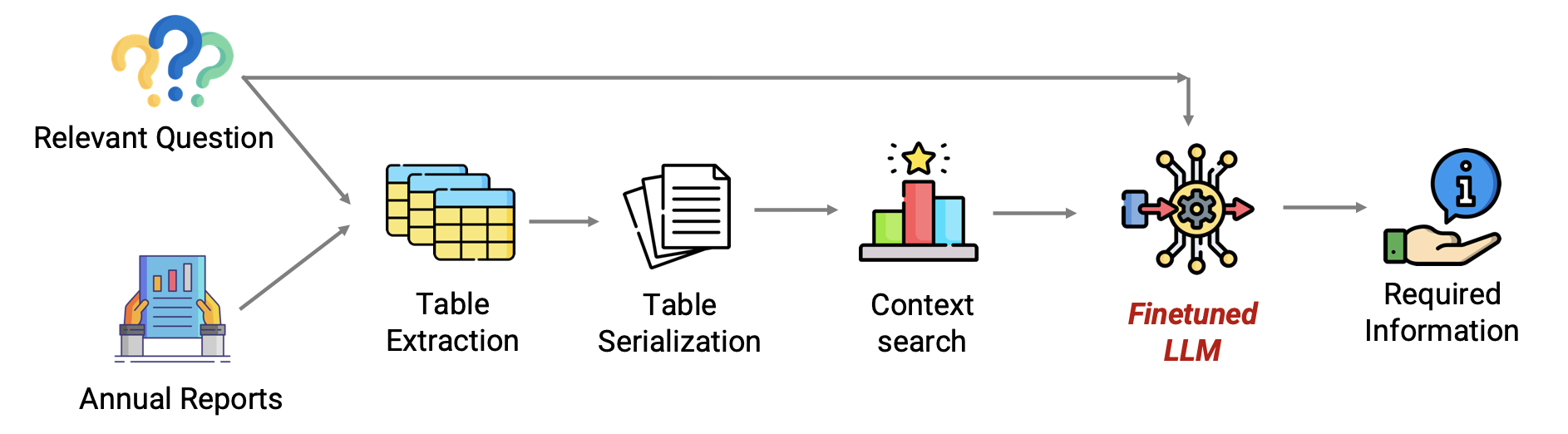}
    \caption{Schematic describing the different steps in our pipeline}
    \label{fig:pipeline}
\end{figure}

\subsection{Table extraction}
In the quest to extract tables from the PDF report, our efforts delved into various approaches, with a primary reliance on OCR techniques operating behind the scenes. These techniques were instrumental in identifying the specific bounded areas housing tables within the document, allowing us to subsequently parse and extract their contents. Among the array of methods tested, Pytabula emerged as the frontrunner, showcasing a notable efficacy in retrieving tabular data accurately.

However, our journey wasn't without hurdles. Pytabula, while efficient in conventional table structures, grappled when faced with atypical formats, especially those incorporating intricate elements like subheadings and subcolumns. The tool's proficiency seemed to dwindle in such scenarios, posing a challenge in extracting data accurately from these non-standard structures. Furthermore, a recurring issue surfaced where tabula occasionally misinterpreted infographics within the document as tables, leading to inaccuracies in the extracted data. These shortcomings served as a catalyst, emphasizing the need for enhancements in this component's functionality to adeptly handle diverse and complex table formats for more precise extraction of information.

\begin{figure}[ht]
    \centering
    \includegraphics[width=0.5\linewidth]{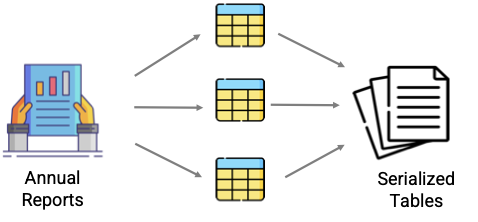}
    \caption{Firstly we extract all the tables from the pdf report and subsequently serialise them}
    \label{fig:table_extraction_serialisation}
\end{figure}

\subsection{Table to Text Serialization}
In our research, we identified the transformation of structured tabular data into cohesive, meaningful text as a crucial component of our processing pipeline. This conversion was established as a significant contributor to the overall performance of our model. Specifically we try two serialisation approaches, with their effect on QA results given in \ref{sec:exp_results}.

\subsubsection{Naive Serialisation}
\label{sec:naive-ser}

\begin{figure}[ht]
\begin{minipage}{0.5\textwidth}
In a naive serialization approach \ref{fig:naive_ser}, leveraging header information is a fundamental step in the process of serializing a table. The header serves as a crucial guide that provides metadata about the structure and characteristics of the table. By intelligently incorporating the header data into the serialization process, we ensure that the serialized output accurately represents the original table's structure. We took inspiration of serialization from \cite{hegselmann2023tabllm}
\end{minipage}
\hfill
\begin{minipage}{0.5\textwidth}
    \centering
    \includegraphics[width=0.7\linewidth]{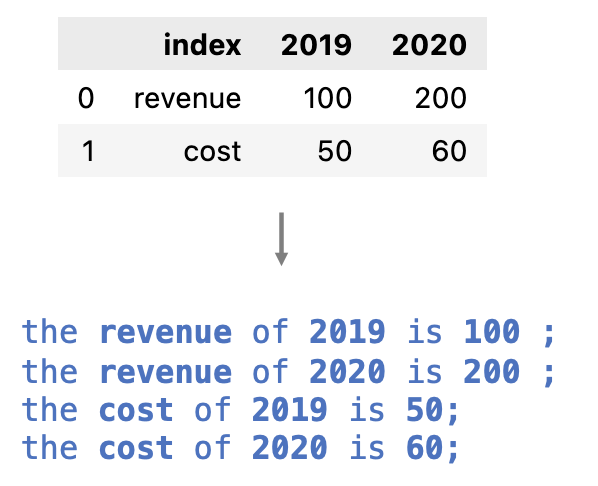}
    \caption{example of naive serialization}
    \label{fig:naive_ser}
\end{minipage}
\end{figure}

\subsubsection{LLM Serialisation}
However, in our experiments we observed that a naive serialization method was sometimes inadequate in effectively handling the complexities inherent in multi-index and non-standard table formats. Such rudimentary approaches not only failed to capture the nuanced intricacies of these tables but also detrimentally impacted the performance of large language models. This was primarily due to the fact that fine-tuning the models on semantically incorrect contexts often led to inaccurate outputs. Using LLMs for serialisation enabled us to create a robust framework for serializing tables into text that was not only syntactically correct but also semantically relevant. Our experiments and results demonstrate the effectiveness of this approach in enhancing the model's ability to interpret and process complex tabular data.

We utilized GPT-3.5 for the initial generation of serializations. Subsequently, we delved into both zero-shot and few-shot prompting techniques, employing the Llama-7b chat model to replicate the data serialization achieved with GPT-3.5. The utilization of the Llama model is intended to facilitate open-source accessibility of our solution. The comparison between Llama serialization and GPT-3.5 answers is presented in Table \ref{tab:rouge-scores}. In alignment with similar literature, we utilize the Rouge score for comparison, as outlined in \cite{banerjee2023benchmarking}.

\begin{table}
  \centering
  \begin{tabular}{lcccc}
    \toprule
    & \textbf{rouge1} & \textbf{rouge2} & \textbf{rougeL} & \textbf{rougeLsum} \\
    \midrule
    \textbf{Zeroshot} & 0.4665 & 0.2286 & 0.3265 & 0.4140 \\
    \textbf{Fewshot} & \textbf{0.5793} & \textbf{0.3476} & \textbf{0.4353} & \textbf{0.4508} \\
    \bottomrule
  \end{tabular}
  \caption{ROUGE scores for Llama serialisation with GPT3.5 benchmark}
  \label{tab:rouge-scores}
\end{table}

\subsection{Context Search}
Segmenting larger contexts into smaller, more digestible chunks has proven to be highly effective, particularly in enhancing the performance of Large Language Models (LLMs) during question answering tasks, ensuring precision and reducing the incidence of erroneous or unrelated outputs (hallucinations). For instance, the Llama-7b model, despite its considerable 4k context length capability, tends to predominantly focus on the recent or latter parts of the context, often neglecting information that may lie in the middle. This selective attention can lead to gaps in information processing. By implementing a chunking strategy, we can provide these LLMs with compact and focused context portions. This practice is beneficial as it ensures that the model considers all relevant information, irrespective of its position in the overall context. Consequently, chunking can significantly enhance an LLM's ability to reason and answer questions more accurately, by mitigating the tendency to overlook critical information embedded within longer text passages.

\begin{figure}[!ht]
\begin{minipage}{0.5\textwidth}
Our approach to pinpointing relevant text chunks involves utilizing the FAISS DB to fetch similarity score, which gauges the similarity between a given question and specific segments of text. Once we extract and serialize tables, we segment the text into smaller pieces. Subsequently, we compute the FAISS score between each text chunk and the question, ultimately selecting the text segment with the highest FAISS score.   
\end{minipage}
\hfill
\begin{minipage}{0.45\textwidth}
    \centering
    \includegraphics[width=1\linewidth]{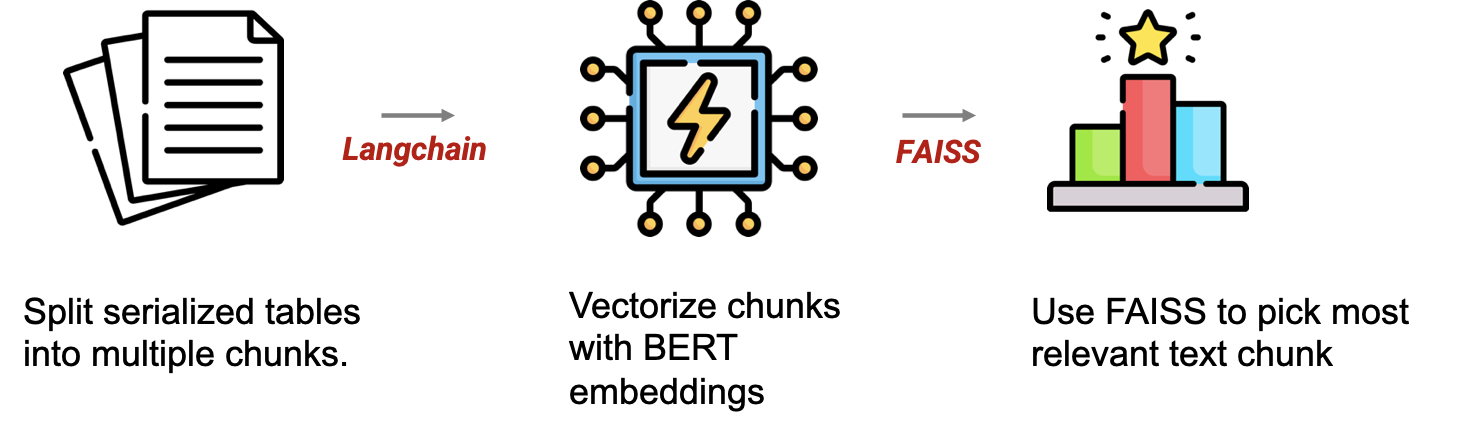}
    \caption{Context chunking and search process with FAISS}
    \label{fig:context_search}
\end{minipage}    
\end{figure}


\subsection{Numerical Question Answering with LLMs}
In the pivotal phase of our project, we focused intensively on evaluating the prowess of Large Language Models (LLMs) in executing numerical reasoning tasks, specifically within the context of financial report tables that were formatted using the previously described serialization method. Recent literature has shown considerable succes in this task, such as works by \cite{kasai2022realtime} and \cite{ngai2021transformerbased}. The primary objective here was to harness the LLMs to accurately extract critical numerical data from these tables and then apply the necessary arithmetic operations to address the given queries. This strategy of decomposing intricate problems into more manageable segments significantly enhances our ability to utilize the advanced text-generation capabilities of LLMs in tackling complex numerical reasoning challenges.

Our experiments were conducted using both the Decoder-only model (Llama 2-7b-chat, Llama 2 7b) and the Encoder-decoder model (T5). Specifically, we engaged the Llama-7b chat model in a limited example setting and also refined its performance through fine-tuning with the QLoRA technique. The T5 model on the other hand was  This approach underscores our commitment to pushing the boundaries of LLM applications in numerical data processing and analysis.
Within our project, our primary focus centers on the tabulated data encapsulated within financial reports. Through our exploration of datasets, we observed a significant dependency of Financial Q\&A on information presented in tables. However, we encountered challenges in extracting and analyzing tabulated data using Language Models (LLMs). Consequently, our project is dedicated to addressing this hurdle by honing in on tabulated data within financial reports, aiming to conduct numerical reasoning and analysis on this specific data format.


\subsection{Post Processing}
\label{sec:post_process}
In our study, addressing numerical reasoning tasks using a language model necessitated an additional step: the transformation of the model's textual output into a more structured and easily parseable format. This conversion facilitates a systematic evaluation of the model's responses, encompassing aspects such as variable computation, the identification of correct arithmetic operations, and the final result determination.
To optimize this process, we meticulously designed our input prompts, both during fine-tuning and few-shot prompting, to ensure they were structured in a way that would allow for efficient application of regular expressions (regex). This approach was instrumental in extracting key elements from the answers generated by the model, enabling a direct comparison with the actual, correct answers.

\begin{figure}[ht]
\begin{minipage}{0.5\textwidth}
Implementing this post-processing step not only allowed us to rigorously evaluate the model using a variety of quantitative metrics but also provided insights into specific areas where the model was underperforming or erring. By pinpointing these discrepancies, we could more accurately determine the necessary adjustments and improvements to enhance the model's performance. This methodical approach to post-processing and analysis forms a critical component of our research, contributing significantly to the overall development and refinement of the model's capabilities in numerical reasoning.
\end{minipage}
\hfill
\begin{minipage}{0.45\textwidth}
    \centering
    \includegraphics[width=1\linewidth]{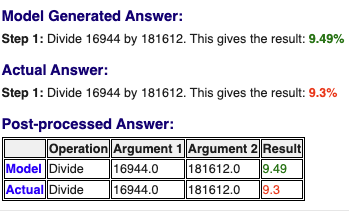}
    \caption{Postprocessing to parse operators and numerical arguments from answers}
    \label{fig:output_table}
\end{minipage}
\end{figure}

\section{Experiments and Results}
\label{sec:exp_results}

\subsection{QA with T5 \& Naive serialization}
In this section, we delve into the results obtained through the application of a fine-tuned T5 model in conjunction with a naive serialization approach as decribed in \ref{sec:naive-ser} for inputs and outputs as shown in \ref{fig:table_extraction_serialisation}. For the finetuning process we used the batch size as 8, the number of epochs as 12, and utilizes the Adam optimizer with a learning rate of 0.0001. The subsequent analysis and outcomes detailed in this section shed light on the effectiveness and performance of this combined approach in generating accurate and contextually relevant results for numerical reasoning queries within the domain of financial reporting.

First, we conduct a comparative analysis between the number of steps generated in the predicted answers and the actual answers, on the validation set of FinQA. The outcomes of this comparison are presented in \ref{tab:res_steps}. While we observe a general alignment between the predicted and true number of steps in the answers, it is noteworthy that the T5 model exhibits a propensity to overcomplicate responses by predicting additional steps. This tendency underscores a nuanced aspect of the model's behavior that warrants careful consideration in assessing the overall accuracy and simplicity of the generated answers.

\begin{table}[ht]
  \centering
  \begin{tabular}{lccc}
    \toprule
    \textbf{steps label vs steps generated} & \textbf{1} & \textbf{2} & \textbf{$>$2} \\
    \midrule
    \textbf{1} & \textbf{355} & 71 & 63 \\
    \textbf{2} & 21 & \textbf{204} & 62 \\
    \textbf{$>$2} & 6 & 19 & \textbf{47} \\
    \bottomrule
  \end{tabular}
  \caption{Crosstab between number of steps in generated text and label text. We see that the T5 model has a tendency to add imaginairy steps}
  \label{tab:res_steps}
\end{table}

Subsequently, our attention shifts to instances where the predicted and true number of steps align. In a detailed examination, we scrutinize the accuracy of the model in correctly identifying both the arguments and the operators essential for generating numerical answers in these cases, as described in post processing \ref{sec:post_process}. This focused analysis aims to provide insights into the model's proficiency in not only capturing the correct number of steps but also in discerning the precise components required for formulating accurate numerical responses. As we can see from the result table, result with calculator is generate results that is significantly better. We're using the same post processing method as mentioned from above section.

\begin{table}[!ht]
  \begin{minipage}{0.4\textwidth}
    \centering
    \includegraphics[width=0.7\linewidth]{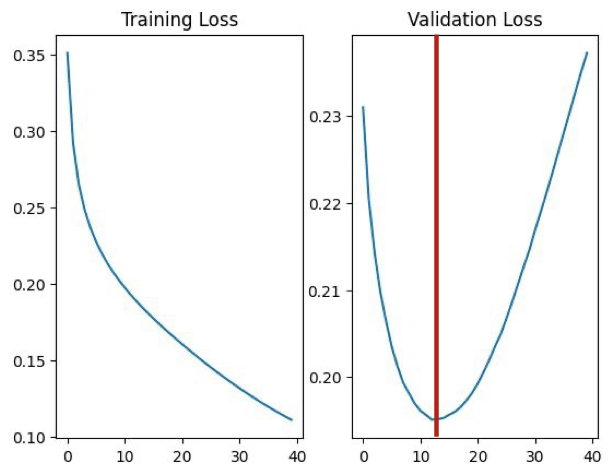}
    \caption{Training Curve for finetuning of T5}
    \label{fig:tf_training_curve}
  \end{minipage}%
  \begin{minipage}{0.4\textwidth}
    \centering
    \begin{tabular}{lr}
      \toprule
      \textbf{Match b/w predicted and generated steps}\\
      \toprule
      \textit{Numerical Operator} & 94.9\% \\
      \textit{Argument 1} & 63.9\% \\
      \textit{Argument 2} & 70.4\% \\
      \textit{Result (w/o calculator)} & 11.8\% \\
      \textit{Result (with calculator)} & 62.3\% \\
      \bottomrule
    \end{tabular}
    \caption{Accuracy of T5 Model for Numerical Components}
    \label{tab:correct_match}
  \end{minipage}
\end{table}

The results indicate that the T5 model accurately extracts numerical operations approximately 95\% of the time. This high accuracy is primarily because the number of operators is limited to a few operations like $\{+,-,/,*,\max,\min\}$. Also, when it comes to extracting numerical values, the performance of T5 is around 65\% accurate, which is also competitive, and reveals that the model is able to track the right values most of the time. However model's ability to perform numerical calculations is comparatively bad, leading to a 10\% accurarcy rate in the final calculations. Nevertheless, through post-processing, we observed a substantial improvement in the accuracy of the final results. Allowing for a 10\% deviation as acceptable, we find that post-processing significantly enhances the accuracy rate of the final output to upwards of 60\%.

\subsection{QA with Llama  \& LLM serialisation}
We utilized two open-source models, Llama-2-7b and Llama-2-7b-chat, as the foundation of our experimental framework. Our methodology was centered around two distinct approaches: instruction-based prompting and fine-tuning of the aforementioned models. The instruction finetuning leveraging quantized LoRA was inspired by \cite{chung2022scaling}, \cite{hu2021LoRA}, \cite{dettmers2023QLoRA}.The instruction-based prompting was implemented in a few-shot learning context, while fine-tuning involved adapting the models more extensively to the specific nuances of the dataset and tasks at hand. This dual approach allowed us to comprehensively assess the capabilities and limitations of the Llama2-7b and Llama-2-7b-chat models in the domain of numerical reasoning within a financial context.\\\\
\textbf{Model Specifications, Finetuning method , Training Arguments}\\
In our finetuning methodology, we enhanced the Llama-2-7b-chat model using the Quantized LoRA technique, aiming for computational efficiency and precision. The model, identified by its Hugging Face hub ID, was configured with 4-bit quantization and 16-bit floating-point precision for computations. The tokenizer was adjusted for right-side padding, aligning with the model's inference patterns.
The model architecture was fine-tuned to target specific projection modules, employing a rank of 16 and a projection dimension of 64, without biases in LoRA layers. Training leveraged SFTTrainer with a batch size of 4 on a single A100 GPU, using mixed precision and a cosine learning rate schedule. The configuration promoted both efficiency and model's numerical reasoning, preparing it for tasks requiring precise computational outputs. 
\\
\begin{figure}[!ht]
  \begin{minipage}{0.5\textwidth}
    \centering
    \includegraphics[width=0.7\linewidth]{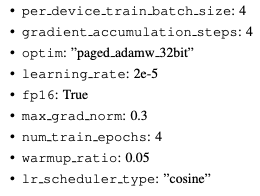}
    \caption{Training Configuration for Optimal Performance}
    \label{fig:Training_Configurations}
  \end{minipage}%
\begin{minipage}{0.5\textwidth}
    \centering
    \includegraphics[width=1\linewidth]{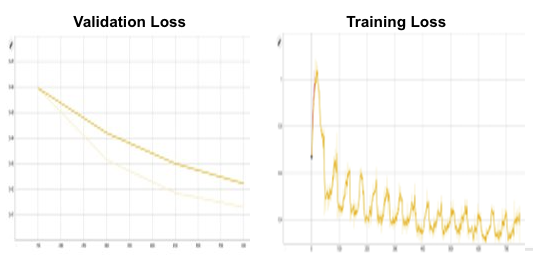}
    \caption{Loss Plots for Finetuned Llama2-7b}
    \label{fig:llama_loss_plots}
  \end{minipage}
\end{figure}
\\
\textbf{Results}\\
The results were evaluated on validation dataset comprising of 256 samples. As discussed, the quantitative metrics were calculated by converting the model generated answer in a parseable structure as shown above in Figure \ref{fig:table_extraction_serialisation}.
Both Results in Table \ref{tab:llama_results_1} and \ref{tab:llama_results_2} are computed using the post-processed output.\\
\\\textbf{Table \ref{tab:llama_results_1}} - EM stands for exact match\ and Results here correspond's to model's output w/o calculator. Arg1 and Arg2 refer to Arguments 1 and 2 respectively.\\
The experimental outcomes demonstrate that the application of the Quantized LoRA technique to fine-tune the Llama-2-7b-chat model produces superior performance across a broad spectrum of metrics. The fine-tuned models demonstrate enhanced congruence with the expected results, as evidenced by improved exact match scores. This enhancement indicates that fine-tuning instills a greater degree of computational accuracy within the outputs of the language model.
The observed mean deviations were notably substantial, primarily attributable to instances involving calculations with large numerical values where the model's outputs were significantly inaccurate.

\begin{table}
  \centering
    \resizebox{\textwidth}{!}{
  \begin{tabular}{lccccc}
    \toprule
    \textbf{Model} & \textbf{Variant} & \textbf{Arg 1 - EM} & \textbf{Arg 2 - EM} & \textbf{Operator - EM} & \textbf{Result - EM} \\
    \midrule
    \textbf{Llama 2-7b} & \textbf{FewShot Prompting} & \text{36.84\%} & \text{40.35\%}  & \text{82.46\%} & \text{7.02\%}  \\
    \textbf{Llama 2-7b} & \textbf{Finetuned - QLoRA} & \text{43.80\%} & \text{45.60\%}  & \text{77.20\%} & \text{17.50\%} \\
    \textbf{Llama 2-7b Chat} & \textbf{FewShot Prompting} & \text{42.80\%} & \text{44.70\%}  & \text{87.50\%} & \text{8.10\%}
    \\
    \textbf{Llama 2-7b Chat} & \textbf{Finetuned - QLoRA} & \textbf{51.40\%} & \textbf{52.70\%}  & \textbf{88.60\%} & \textbf{20.00\%}
    \\
    \bottomrule
  \end{tabular}
  }
  \caption{Exact Match scores for various components involved in numerical reasoning as \%'s}
  \label{tab:llama_results_1}
\end{table}

\begin{table}
  \centering
  \resizebox{\textwidth}{!}{
  \begin{tabular}{lccccc}
    \toprule
    \textbf{Model} & \textbf{Variant} & \textbf{Result Deviation} & \textbf{Computed Result Deviation} & \textbf{RougeL Score}\\
    \midrule
    \textbf{Llama 2-7b} & \textbf{FewShot Prompting} & \text{$>$100k} & \text{8.557}  & \text{0.665}  \\
    \textbf{Llama 2-7b} & \textbf{Finetuned - QLoRA} & \text{$>$100k} & \text{5.099}  & \text{0.688}  \\
    \textbf{Llama 2-7b Chat} & \textbf{FewShot Prompting} & \text{91871.71} & \text{12.771}  & \text{0.637} 
    \\
    \textbf{Llama 2-7b Chat} & \textbf{Finetuned - QLoRA} & \textbf{17451.678} & \textbf{9.084}  & \textbf{0.714} 
    \\
    \bottomrule
  \end{tabular}
  }
  \caption{Mean Result deviation and the RougeL scores computed across all variants}
  \label{tab:llama_results_2}
\end{table}

\textbf{Table \ref{tab:llama_results_2}} - Results show the deviation between model generated results and computed results (using python interpreter) wrt to actual results. RougeL score is computed between model's text output vs original answers.

The findings presented herein illustrate the discrepancies between outcomes generated by language models and those derived via a Python interpreter, in comparison to actual results. Notably, the data underscores a fundamental challenge: while language models can accurately identify appropriate argument values and operators, they often falter in executing the final computation accurately. This shortcoming underscores the potential benefits of adopting Program Aided Language Models (PAL)\cite{gao2023pal} that integrate traditional programming capabilities within a language model framework. To address this, we incorporated a Python interpreter for result computation, which, despite yielding a closer approximation, still manifested minor discrepancies. These variations suggest the presence of underlying deficiencies within the input dataset, indicating areas for future refinement.

The results also demonstrate the utility of fine-tuning the Llama2 7b chat model using QLoRA as it gave the best results across all the metrics. For the Llama 2-7b model, the result deviations are capped at 100K as they were higher.

\subsection{Qualitative Analysis of Discrepancies}
The following figure \ref{fig:observable_discrepancies} provides an overview of prevalent challenges encountered during the response generation process. The observed inconsistencies highlight that refinements in prompt engineering and enhancements in model architecture have the potential to mitigate these issues.\\
\begin{figure}[ht]
    \centering
    \includegraphics[width=1\linewidth]{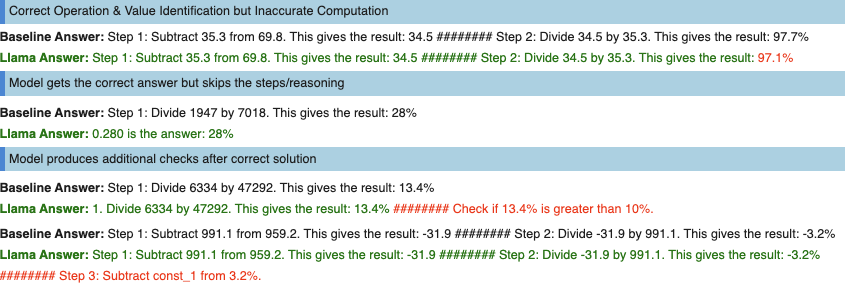}
    \caption{Overview of Common Discrepancies Observed in Model's Response}
    \label{fig:observable_discrepancies}
\end{figure}


\section{Conclusion}
As part of this exercise we have been able to engineer a robust approach for numerical question answering from PDF reports by leveraging advanced natural language processing techniques, based on T5, Llama-2 and Langchain.
Despite the overall success, certain nuances required careful consideration. Parsing non-conventional tables presented challenges, as the approach encountered limitations in handling unconventional table structures commonly found in complex reports. Additionally, errors in the table-to-text serialization process posed another obstacle, demanding a refined post-processing pipeline to enhance the accuracy of the answers generated. Furthermore, it was observed that certain aspects of question-answering with large language models (LLMs), exhibited subpar performance, necessitating ongoing efforts to address and improve model performance.

In conclusion, while our approach demonstrated commendable results in numerical question answering from PDF reports, continual refinement is essential to overcome challenges related to non-conventional tables, serialization errors, and performance nuances associated with large language models in question-answering tasks. These insights guide our ongoing efforts to enhance the robustness and applicability of our approach.

\section{FUTURE WORK}
Based on the results of the current research, the team plans to enhance the model's capability in parsing and serializing complex table structures in financial reports. They also aim to address the challenges identified in handling large numerical values and refine the dataset to improve overall accuracy. Future efforts will involve integrating Program-Aided Language Models (PALs) to further advance numerical reasoning capabilities in the domain of financial analysis.
\bibliography{iclr2021_conference}
\bibliographystyle{iclr2021_conference}

\end{document}